\def\year{2019}\relax
\pdfoutput=1
\documentclass[letterpaper]{article} 
\usepackage{aaai19}  
\usepackage{times}  
\usepackage{helvet} 
\usepackage{courier}  
\usepackage[hyphens]{url}  
\usepackage{graphicx} 
\usepackage{graphicx}  
\usepackage{amsmath}
\usepackage{amssymb}
\usepackage[acronym]{glossaries}
\usepackage{listings}
\usepackage{color}
\usepackage{hyperref}
\usepackage{flushend}

\definecolor{codegreen}{rgb}{0,0.6,0}
\definecolor{codegray}{rgb}{0.5,0.5,0.5}
\definecolor{codepurple}{rgb}{0.58,0,0.82}
\definecolor{backcolour}{rgb}{0.95,0.95,0.92}

\lstdefinestyle{mystyle}{
    backgroundcolor=\color{backcolour},   
    commentstyle=\color{codegreen},
    keywordstyle=\color{magenta},
    numberstyle=\tiny\color{codegray},
    stringstyle=\color{codepurple},
    basicstyle=\footnotesize,
    breakatwhitespace=false,         
    breaklines=true,                 
    captionpos=b,                    
    keepspaces=true,                 
    numbers=left,                    
    numbersep=5pt,                  
    showspaces=false,                
    showstringspaces=false,
    showtabs=false,                  
    tabsize=2
}

\title{Petri Net Machines for Human-Agent Interaction}
\author{Christian Dondrup, Ioannis Papaioannou, Oliver Lemon\\ 
School of Mathematical and Computer Sciences, Heriot-Watt University\\ 
Edinburgh Campus, Edinburgh, EH14 4AS\\
\{c.dondrup, i.papaioannou, o.lemon\}@hw.ac.uk 
}

\newacronym{kb}{KB}{Knowledge-Base}
\newacronym{dfsm}{DFSM}{Deterministic Finite State Machine}
\newacronym{fsm}{FSM}{Finite State Machine}
\newacronym{pn}{PN}{Petri-Net}
\newacronym{pnm}{PNM}{Petri-Net Machine}
\newacronym{ros}{ROS}{Robot Operating System}
\newacronym{hai}{HAI}{Human-Agent Interaction}
\newacronym{hri}{HRI}{Human-Robot Interaction}
\newacronym{sds}{SDS}{Spoken Dialogue Systems}
\newacronym{dqn}{DQN}{Deep Q-Network}
\newacronym{rl}{RL}{Reinforcement Learning}
\newacronym{pddl}{PDDL}{Planning Domain Definition Language}
\newacronym{smach}{SMACH}{State MACHine}
\newacronym{pnp}{PNP}{Petri Net Plan}
\newacronym{yaml}{YAML}{Yet Another Markup Language}

\nocopyright

\begin{document}

\maketitle

\begin{abstract}
Smart speakers and robots become ever more prevalent in our daily lives. These agents are able to execute a wide range of tasks and actions and, therefore, need systems to control their execution. Current state-of-the-art such as (deep) reinforcement learning, however, requires vast amounts of data for training which is often hard to come by when interacting with humans. To overcome this issue, most systems still rely on Finite State Machines. We introduce Petri Net Machines which present a formal definition for state machines based on Petri Nets that are able to execute concurrent actions reliably, execute and interleave several plans at the same time, and provide an easy to use modelling language. We show their workings based on the example of Human-Robot Interaction in a shopping mall.
\end{abstract}

\section{Introduction}
\label{sec:intro}

Smart devices such as the Google Home or Amazon Echo, and even social robots such as Anki's Cozmo and SoftBank's Pepper have moved from the lab into private homes. All of these agents seek to interact with humans in their vicinity in one way or the other. Hence, all of them need a way to manage this interaction and a lot of approaches to solve this problem have been proposed over the years. On the one hand, deep reinforcement learning such as \glspl{dqn} have been used for many applications with one of them being behaviour generation via the learning of control policies, e.g. playing games~\cite{mnih2015human}. This, however, requires access to large amounts of meaningful raw sensor data which is difficult to come by when looking at the domain of \gls{hai} in general or \gls{hri} in particular. Traditional reinforcement learning approaches such as Q-Learning have similar problems or require a long time exploring the state space to create a suitable policy. Exploration, however, can rarely be done in \gls{hai} scenarios and is, therefore, often scaffolded by simulating user interactions. This is particularly prevalent in \gls{sds}, see surveys by e.g.~\cite{schatzmann2006survey,pietquin2013survey}. Using a \gls{fsm} on the other hand allows to model interaction using expert knowledge and creating behaviours for interaction without needing to collect data a-priori and/or to first create a user simulation. For this reason, many interactive systems in \gls{hri} and for dialogue management for smart speakers still rely on \glspl{fsm}. Examples for this are \gls{smach}~\cite{bohren2010smach} for the \gls{ros} or \cite{thomson2010bayesian,curryalana} in the wider sense in case of \gls{sds}.

\begin{figure}[t]
  \centering
  \includegraphics[width=.65\linewidth]{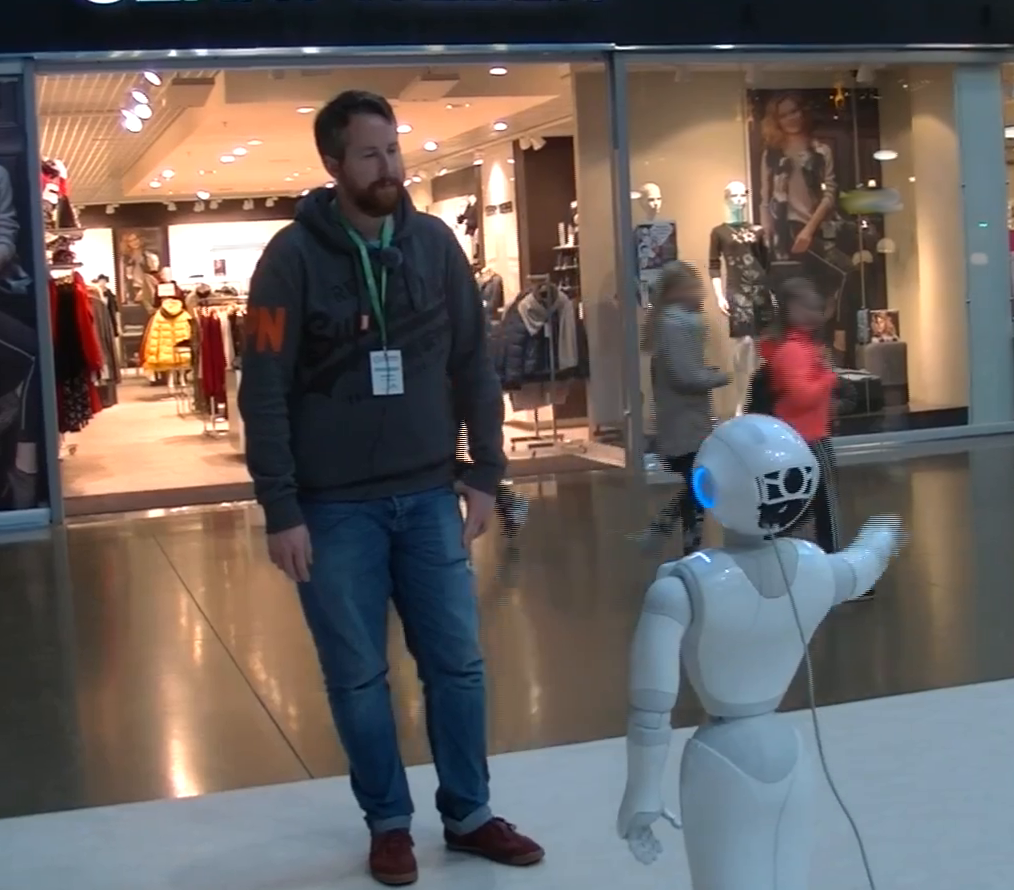}
  \caption{One of the experimenters interacting with the system embodied by the robot. Shot on location in a Finnish shopping mall.}
  \label{fig:ideapark_me}
\end{figure}

Another challenge when it comes to \gls{hai} and \gls{hri} is that a myriad of actions might have to happen at the same time. An example for this in \gls{hri} would be a robot giving a route description to an interaction partner. In this case the robot has to verbalise the description, point to the target, look at the human or the direction pointed to, etc. Modelling these concurrent actions correctly in a \gls{fsm} presents problems of managing these actions and the general flow of the interaction due to its exploding complexity.
In this paper, we present an approach to building \glspl{dfsm} automatically. These \glspl{dfsm} are based on \glspl{pn} which are able to handle concurrency and guarantee that there are no unreachable states. This approach was inspired by work by \cite{ziparo2011petri} but we believe that it improves not only the implementation and usage but also the conceptual realisation of the \gls{pn} making this approach even more versatile and powerful. 
Moreover, the approach presented here also allows to execute several \glspl{pn} concurrently to allow to interleave tasks or execute them at the same time.

In this paper we present our execution system based on \glspl{pn} developed with the \acrfull{ros} in mind. It natively supports \gls{ros} action servers and presents an alternative to \gls{smach}~\cite{bohren2010smach}. It's current application is to create a unified plan modelling and execution framework combining dialogue and physical actions on a robot. Most current dialogue systems apart from the one this execution system is part of are either task or social chat based but not both. Therefore, in previous work, we argued why it would be beneficial to combine the two \cite{papaioannou2017hybrid}. While this was the main initial motivation, the resulting execution system is more generally applicable and can be used for all tasks one would use a \gls{fsm} for. However, it comes with all the benefits of a \gls{pn} such as concurrent execution, modelling complex structures, and in addition to that as we believe an easy to use description language to create state machines without having to do much programming. We tested the system in a deployment of the robot in a Finnish shopping mall (see Figure~\ref{fig:ideapark_me}) where it gave directions to customers.

\section{Related Work}

There are many approaches when it comes to selecting and executing actions on robots. Some of the more notable ones are \cite{ingrand1996prs,kim2001executing,verma2005plan,lesire2018aspic,colledanchise2018improving,xu2002modeling,king2003coordinated}. All of these approaches have in common that they are designed for the planning and execution of sequences of actions which are disjoint meaning that the output of one action is not used by a different action further down the sequence. This, however, is possible using the approach presented here. Additionally, the system we propose also offers communication with an external \gls{kb} which can even be a user of a dialogue system to fill gaps before or during execution of an action. Moreover, the presented system automatically generates checks and recovery behaviours based on preconditions and effects of each action to make execution more stable. 

In the following, we present more detail on the two concepts most closely related to the work presented here, i.e. \glspl{fsm} and \glspl{pn}. Another popular approach, i.e. \gls{rl}, has the draw back of either having to explore the state space or needing vast amounts of data collected a-priori. Hence, we will not go into detail about approaches using \gls{rl} for agent control.
\vspace{-.5cm}

\paragraph*{Finite State Machines}
One can implement a \gls{fsm} in any programming language. There are some examples where these \glspl{fsm} have been specialised for robot control. An early example of using \glspl{fsm} for robot control is the work by \cite{brooks1986robust} to control a mobile robot and give it some degree of autonomy. The system was divided into several smaller modules which were implemented as \glspl{fsm} and communicated with each other via a network. As mentioned earlier, the most notable due to the widely used \gls{ros} is \gls{smach}~\cite{bohren2010smach}. It was created for rapid development and provides convenient introspection tools. Instead of creating \glspl{fsm} by hand, others aim to create them automatically such as ROSPlan~\cite{cashmore2015rosplan}. Here, the authors of the paper use a \gls{pddl} based planner to create a sequence of actions to be executed to reach a given goal state from the current start state. ROSPlan also provides a \gls{ros} interface to execute those actions in the order given by the planner. This sequence of actions itself can also be regarded as a \gls{fsm} in the wider sense.

The vast majority of \gls{fsm} based approaches suffer from the same short comings, i.e. the need to model the flow manually and the probability of introducing errors during modelling, the difficulty modelling concurrent states on one or multiple agents, and the problem of not being able to deal with environment states that have not been modelled. Our approach presented in this paper aims to deal with some of these issues by using \glspl{pn} to model concurrency and guarantee that all states are reachable, and by using a simple modelling language. It also provides checks and recovery behaviours to deal with unexpected environment states.
\vspace{-.2cm}

\paragraph*{Petri Nets for robot control}
In the past, there has already been work on modelling robot behaviour as \glspl{pn}. Some examples of modelling (multi-agent) robotic systems are \cite{milutinovic2002petri,sheng2005peer,costelha2007modelling}, where \cite{costelha2007modelling} also explicitly include a model of the environment. Most of the systems that used \glspl{pn} in robotics, however, are either modelling ad-hoc solutions to specific problems or used 3rd-party methods for task execution. The first approach of defining a language for so-called \glspl{pnp} was presented in \cite{ziparo2011petri}. This approach has later on been used in different scenarios of \gls{hri} such as the generation of social-plans, i.e. plans that include both actions of robot and human ~\cite{nardi2014representation}, and the explicit inclusion of social norms in \glspl{pnp}~\cite{carlucci2015explicit}. For this reason, the work presented here builds on \glspl{pnp}. In previous work~\cite{dondrup2017introducing}, we already extended \glspl{pnp} to work in combination with ROSPlan~\cite{cashmore2015rosplan} to automatically translate the plan created into a \gls{pnp} for execution using the defined preconditions and effects to automatically generate checks and recovery behaviours. A similar system without the use of recovery behaviours generated based on the plan has also been presented by \cite{sanelli2017short}.

The work presented in the following builds on the concepts by \cite{ziparo2011petri} and our previous work \cite{dondrup2017introducing} and extends its functionality to more closely resemble that of a \gls{pn}. Moreover, it defines a modelling language, supports automated \gls{pn} generation, has tight \gls{ros} integration, allows the concurrent execution of multiple \glspl{pn}, and provides an improved implementation.

\section{Petri Nets}

\begin{figure*}[t]
  \centering
  \includegraphics[width=.8\linewidth]{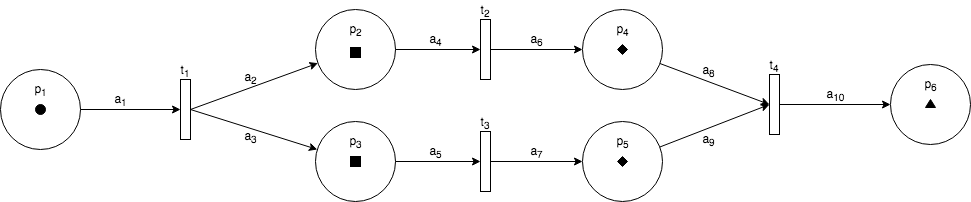}
  \caption{Example of concurrent states in a \gls{pn}. The initial marking $m_i=\{p_1\}$ is shown by dots, squares show $m_{i+1}=\{p_2,p_3\}$, the diamonds show $m_{i+2}=\{p_4,p_5\}$, and the triangle shows $m_{i+3}=\{p_6\}$. All arcs have a weight of $1$ which according to convention has been omitted. $t_1$ is referred to as a fork and $t_4$ as a join.}
  \label{fig:petri_net_concurrent}
\end{figure*}

\begin{figure}[t]
  \centering
  \includegraphics[width=.8\linewidth]{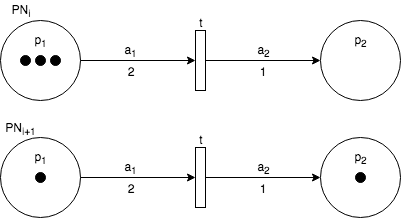}
  \caption{Simple \gls{pn} example. $PN_i$ (top) shows a \gls{pn} with marking $m_i=\{p_1,p_1,p_1\}$ and weights $W=\{a_1,a_1,a_2\}$. $PN_{i+1}$ (bottom) shows the \gls{pn} after transition $t$ occurred where the marking changes to $m_{i+1}=\{p_1,p_2\}$.}
  \label{fig:petri_net_example}
\end{figure}

Based on work by Petri~\cite{petri1962kommunikation}, the theory, notation, and representation of \glspl{pn} and how they could be applied to modelling and analysing systems of concurrent processes was first introduced by Holt et al.~\cite{holt1968final,holt1970events}. In short, \glspl{pn} are state-transition systems which allow to model automata with concurrent and asynchronous states. In order to define a Petri Net $PN$, we first have to define a net $N=(P,T,A)$ where $P$ and $T$ are disjoint finite sets of places and transitions and $A$ is a set of arcs such that $A\subset(P\times T)\cup(T\times P)$. Given a net $N$, we define a configuration $C$ such that $C\subseteq P$. Both $N$ and $C$ form a so-called \textit{elementary net} $EN=(N,C)$. The resulting definition of a \gls{pn} is

\begin{equation}
    PN=(N,M,W)
\end{equation}

\noindent where $M$ is a multiset of places $P$ and the so-called \textit{marking} of the net $N$ which replaces the configuration $C$. $W$ is the multiset of arcs $A$ so that the count of each arc is a measure of its \textit{weight}. 

Figure~\ref{fig:petri_net_example} shows an example \gls{pn}. 
The tokens (black dots) represent the current marking $m_i\in M$ of the net. $PN_i$ shows the \gls{pn} before transition $t$ occurs and $PN_{i+1}$ shows the same \gls{pn} after the transition. The weight of the arc $a_1$ symbolises the number of tokens required in $p_1$ for transition $t$ to occur. The weight of arc $a_2$ shows the number of tokens placed in $p_2$ after transition $t$ occurred. Hence, one can think of arcs pointing to transitions as consuming an amount of tokens equal to their weight and arcs pointing to places generate an amount of tokens equal to their weight. These two processes are disjoint, meaning that the amount of tokens generated does not correspond to the amount of tokens consumed and vice-versa. Transitions `fire' as often as the amount of tokens allows, meaning that if the marking of $p_1$ is a multiple $n$ of the weight of $a_1$, $t$ fires $n$ times.

The process of changing the marking of $PN$ can easily be calculated using simple matrices~\cite{chen}. Firstly, we define the two $(|T|\times |P|)$ matrices $D-=d_{ij}\in\{0,1\}$ with $d_{ij}=1$ if transition $i$ has input from place $j$ for outgoing arcs and $D+=d_{ij}\in\{0,1\}$ with $d_{ij}=1$ if place $j$ has input from transition $i$ for incoming arcs. The composite change matrix $D$ is then defined as $D=(D+) - (D-)$. The new marking $m_{i+1} = [1,1]$ for $PN_{i+1}$ in Figure~\ref{fig:petri_net_example} can then be calculated from marking $m_i=[3,0]$ in $PN_i$ as follows

\begin{equation}\label{equ:marking_update}
    m_{i+1}=\bar{T_i}D + m_i
\end{equation}

\noindent where $\bar{T_i}$ is a $(1\times |T|)$ matrix with $t_j\in \mathbb{N}_0$ representing the number of times each transition $t_j\in T$ should fire.

From these definitions follows that $\forall p\in P: |A_{in}|\in\{0,1\} \wedge |A_{out}|\in\{0,1\}$ whereas $\forall t\in T: |A_{in}|\in\mathbb{N} \wedge |A_{out}|\in\mathbb{N}$. This fact allows for the modelling of concurrent states mentioned earlier by creating forks and joins using transitions. Figure~\ref{fig:petri_net_concurrent} shows the most simple example of concurrency in \glspl{pn}. The \textit{fork} $t_1$ could theoretically split the execution token into any number of places followed by any number of transitions and places. The \textit{join} $t_4$ could similarly join any number of execution tokens. This mechanism allows to start concurrent processes using the fork and wait for their completion using the join.

\subsection{Petri Net Machines}

It is easy to see the similarities between \glspl{pn} and \glspl{dfsm}. Looking at the definition of $DFSM=(\Sigma,S,s_0,\delta,F)$ with $\Sigma$ being a finite non-empty alphabet of input symbols, $S$ being the set of states, the start state $s_0\in S$, $\delta$ representing the state-transition function $\delta:S\times\Sigma\rightarrow S$, and $F$ being the final or goal state, we can clearly see the similarities. States relate to places and markings, transitions relate to $\delta$, and the initial marking $m_0$ being $s_0$. Only the input alphabet and the goal states are missing. This similarity is not surprising and \glspl{pn} have been used to model and analyse all kinds of automata including \glspl{fsm}.

In this work, we present an approach to not only model \glspl{dfsm} using \glspl{pn} but also present a solution and software framework that incorporates external input to create a transition function 

\begin{equation}
\hat{\delta}:\Sigma\times M\times T\rightarrow \bar{T}
\end{equation}

\noindent which allows us to create our new marking using Equation~\ref{equ:marking_update}. Since \glspl{pn} are a simple but powerful model to describe automata that also deal with concurrency, others have presented similar work, e.g. \cite{ziparo2011petri}. The approach presented here, however, uses the full capabilities of the \gls{pn} model to automatically create and execute \glspl{dfsm} and not just an elementary net $EN$ such as in prior approaches. To this end we define \glspl{pnm} $PNM=(\Sigma, PN, m_0, \hat{\delta}, P_g)$ where $PN$ is a Petri Net and $m_0$ is the initial marking of the net. $P_g\subseteq P$ similar to the $F$ of \glspl{dfsm} represents goal states. Reaching a goal state in a \gls{pnm} translates into a goal place $p_i\in P_g$ being included in the current marking $m_j\in M$. Hence, the execution has reached its goal if $P_g \cap m_j \neq \emptyset$.

\section{ROS Petri Net Machines}

The \acrfull{ros} is a popular choice for a large number of research institutes and the industry when it comes to controlling their robots\footnote{While \gls{ros} is popular in academic research, industrial robotics, and self-driving vehicles, it is almost never used for social robots in real-world products and deployments with the exception of the experiment described blow.}. It has a very modular structure and allows to easily port one's work from one robot to another. Moreover, it is supported by a vast community and offers a great number of state-of-the-art, open-source, and off-the-shelve software components readily available. \gls{ros} also comes with its very own implementation of a state machine, i.e. SMACH~\cite{bohren2010smach}. \gls{smach} is a python framework for creating \glspl{dfsm} but as mentioned earlier suffers from the same short-coming as most implementations of DFSMs. For this reason, we chose to base our implementation of a \gls{dfsm} on \glspl{pnm}. Similar to \cite{ziparo2011petri}, we use \gls{ros} action servers which are triggered whenever a transition occurs, but we also allow to generate the \glspl{pnm} on-the-fly based on prior work~\cite{dondrup2017introducing}.

The system presented in the following is based on \glspl{pnp} by \cite{ziparo2011petri} but extends on their idea and improves the implementation to support automatic generation of \glspl{pnm}, handle concurrent execution of multiple \glspl{pnm}, speed up the execution, and to better exploit the native \gls{ros} infrastructure. The latter is achieved by using \gls{ros} action servers identified via their signature as described below. This aims at making its use easier while increasing its capacity for automation and allow the use of the full modelling prowess of \glspl{pn}. In addition, we also provide a modelling language that makes it easy to define \glspl{pnm} without much programming. The code is open source and freely available.\footnote{\label{foot:link}\url{https://github.com/cdondrup/petri_net/tree/ros}}

\subsection{Action Servers}

\begin{figure}[t]
  \centering
  \includegraphics[width=.7\linewidth]{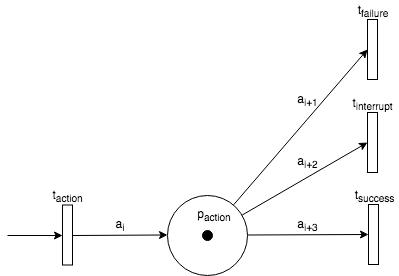}
  \caption{Example of a \gls{ros} action with the 3 outcomes of a typical \gls{ros} action server. When $t_{action}$ is triggered, the action server is started. $\bar{T}$ depends on the reported outcome of the server and only allows the corresponding transition to fire.}
  \label{fig:petri_net_action}
\end{figure}

\textit{Action Servers}\footnote{\url{http://wiki.ros.org/actionlib}} are one of the most used principles in \gls{ros} to execute behaviours on robots. They allow to start, monitor, and interrupt processes remotely. Most actions or behaviours robots can execute are implemented as such action servers. One example being \texttt{move\_base}\footnote{\url{http://wiki.ros.org/move_base}} which has the robot navigate to a given goal. For this reason it makes sense to support this programming principle in any state machine for \gls{ros}. \gls{smach}~\cite{bohren2010smach} for example, offers functionality to use (3rd-party) stand-alone action servers as states directly. We also followed this example and allow to use action servers as places directly which is a novel concepts using \gls{pn}-based execution. These actions are triggered by a transition and while they are executing, the marking reflects this by having a token in the place after the transition that started the server. Once the server has finished, the transitions following the place will become active. Which of these transitions are active depends on how the server finished (see Figure~\ref{fig:petri_net_action}), thereby, creating our transition function from Equation~\ref{equ:marking_update}.

Sometimes, it is necessary for an action to be able to communicate with the process that has started it. For this reason, we implemented a dedicated version of the \gls{ros} action server that is able to query the underlying \gls{kb}. It inherits the same functionality as an action server but also offers query and inform methods (see Listing~\ref{code:kb}). Both types of servers can be used interchangeably.

\noindent\begin{minipage}{\linewidth}
\begin{lstlisting}[language=python, caption={Methods to interact with the \gls{kb}. Both QUERY and UPDATE provide the functionality of specifying the LOCAL, GLOBAL, or ALL \glspl{kb}. The letter first querying LOCAL and if no value is returned the GLOBAL \gls{kb}.}, label=code:kb]
update_kb(RPNActionServer.UPDATE_LOCAL, "spam", "eggs")
query_kb(RPNActionServer.QUERY_ALL, "spam")
\end{lstlisting}
\end{minipage}

\subsection{Knowledge-Base}

One of the problems of state machines is the passing of knowledge between states. While action servers have goal messages that contain data and result messages that contain produced data, planning and execution systems such as ROSPlan~\cite{cashmore2015rosplan} use a database to pass information between states that is not part of the planning domain. Hence, it relies on action $a_i$ to put information in that can subsequently be queried by action $a_{i+1}$. This requires both actions to be specifically designed to be executed in sequence. For our execution framework, we use a local \acrfull{kb} that fills all data fields in the goal of the action to be sent to the server automatically and is updated by the resulting data produced by an action server after its execution finished. These fields to be filled are identified by their name. So if action $a_i$ produces data for variable $v$, it will automatically be used to fill variable $v$ in any followup action. If the two variables do not have the same name, a simple \gls{kb} operation can be included in the plan to save the data under the required name before $a_{i+1}$ is executed\footnote{\gls{kb} operations do not require an action server to be implemented but are executed by the \gls{pnm} directly operating on the \gls{kb}.}. This allows the user to use off-the-shelve (3rd-party) action servers without having to change the names of goal or result parameters and recompile them. Moreover, by creating a disjoint instance of the local \gls{kb} for each \gls{pnm} that is executed, we allow the concurrent execution of several plans (see the \nameref{sec:concurrent} section) each having their own \gls{kb} making it thread safe.

The framework can also be interfaced with a number of global \gls{kb} like a data centre to query information or a dialogue system \cite{papaioannou2017alana,curryalana} such as introduced in \cite{dondrup2017introducing}. This can be used to query information from something like an ontology which is not in the local \gls{kb} or to direct questions to the user of a dialogue system for clarification. Imagine, for example, a system for route guidance that was asked to give directions to a restaurant. There might be several restaurants close by so the \gls{pnm}, after finding all possible instances of restaurants, can query the global \gls{kb} (in this case the user) which restaurant they want to go to before the action of generating a route is executed.

\subsection{Recovery Behaviours}

As presented in previous work \cite{dondrup2017introducing}, before executing an action and after its execution, certain behaviours are automatically inserted to check conditions and recover from execution errors. Before the start of an action, the local database is queried to check if the variables required to fill the goal message are present. If not, the \gls{pnm} reports a failure. After the execution of an action, as can be seen in Figure~\ref{fig:petri_net_action}, sever outcomes are checked. If the execution reported success the \gls{pnm} continues as expected. If the action failed, the \gls{pnm} reports a failure and if the action is interrupted (i.e. preempted in \gls{ros} terminology), followup recoveries can be defined manually or the \gls{pnm} continues regardless. Hence, all these recovery behaviours for action, checks, and recovery behaviours for checks, can be used as is but can also be defined manually. This ranges from simple predefined commands such a retrying an action to defining a whole chain of alternative actions.

\begin{lstlisting}[float=t, caption={Example domain excerpt following a \gls{pddl} inspired syntax including params, preconditions, and effects. The super types \texttt{rpn\_action} and \texttt{ros\_action} define which specific python source files to use and have been omitted.}, label=code:domain]
...
actions:
 dummy_server:
  <<: *rpn_action
  params:
   - "value"
  effects:
   and:
    - Comparison: ["eq", [Query: "time", Query: "value"]]
    - not:
     Comparison: ["ne", [Query: "time", Query: "value"]]
 wait:
  <<: *ros_action
  params:
   - "time"
  preconditions:
   Exists: [Query: "time"]
...
\end{lstlisting}

\subsection{Planning Interactions}

In order to plan interactions, the user can define domains and plans manually. These are then translated into action servers themselves which when started execute the plan as a \gls{pnm}. An example domain can be seen in Listing~\ref{code:domain}. The markup language used is \gls{yaml} and the layout for the domain file is based on \gls{pddl}. Actions are defined with a name, a list of parameters, preconditions, and effects. Preconditions and effects can use logical operations as can be seen from the example in lines 9 -- 11, 17. \texttt{Query} takes a single argument, queries first the local and if the value is not found, the global \gls{kb} for the required information, and returns the value. \texttt{Comparison} takes two arguments: i) the comparison operation, e.g. \texttt{eq} for equals, and ii) a list of arguments to execute the comparison on. \texttt{Exists} checks if the given query returned a result or not.

An example plan can be seen in Listing~\ref{code:plan}. It defines initial knowledge to populate the local \gls{kb} on start-up and the list of actions. The actions are executed in order and can be given arguments for the parameters they expect. If any of the parameters is omitted, it is filled from the \gls{kb} automatically. \texttt{dummy\_action} takes ``value'' as a parameter and returns a ``time''. This ``time'' is then used by the \texttt{wait} actions on lines 6 and 7. \texttt{concurrent\_actions} defines a list of actions that should be executed concurrently. These can also be nested. \texttt{dummy\_server} is defined as a \gls{ros} \gls{pn} action or \texttt{rpn\_action} (see Listing~\ref{code:domain}). This means that it is able to communicate with the \gls{pn} execution server to query or update information at runtime (see Listing~\ref{code:kb}).

\begin{figure}[t]
  \centering
  \includegraphics[width=.95\linewidth]{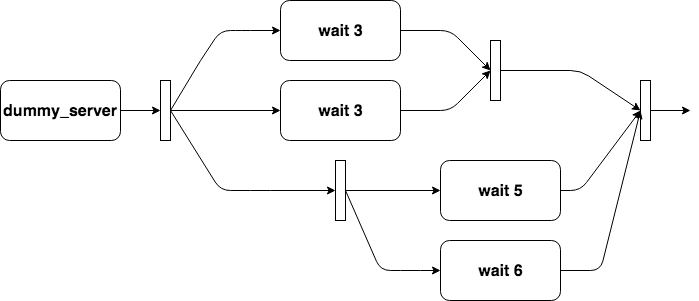}
  \caption{The conceptual structure of the \gls{pnm} resulting from Listing~\ref{code:plan}. Each box contains several places and transitions for precondition and effect checks, and for the execution of the action (see Figure~\ref{fig:petri_net_action}).}
  \label{fig:pnp_plan}
\end{figure}

\begin{lstlisting}[float=t,caption={Example plan. This plan defines initial knowledge that is used to populate the \gls{kb} on start. The plan itself lists (concurrent) actions in order of execution. Line 9 and 10 show how to pass explicit values for parameters where as lines 5 and 6 have the \texttt{wait} function use whatever value is found in the local \gls{kb}. Figure~\ref{fig:pnp_plan} shows the resulting conceptual representation of a \gls{pnm}.}, label=code:plan]
initial_knowledge:
 value: 3
plan:
 - dummy_server: {}
 - concurrent_actions:
  - wait: {}
  - wait: {}
  - concurrent_actions:
   - wait: {time: 5}
   - wait: {time: 6}
\end{lstlisting}

\noindent As can be seen in Figure~\ref{fig:pnp_plan}, the resulting net executes first the \texttt{dummy\_server} and then 4 instances of the \texttt{wait} action concurrently. In addition to concurrent interactions, these plans also allow to create loops and thanks to the mentioned recovery behaviours if-then-else constructs. Neither of these have been shown here due to the limitation of space. The full documentation can be found online\footnotemark[2]. Once a plan has been loaded, it becomes a \gls{ros} action server itself for ease of use.

\begin{figure}[t]
  \centering
  \includegraphics[width=.95\linewidth]{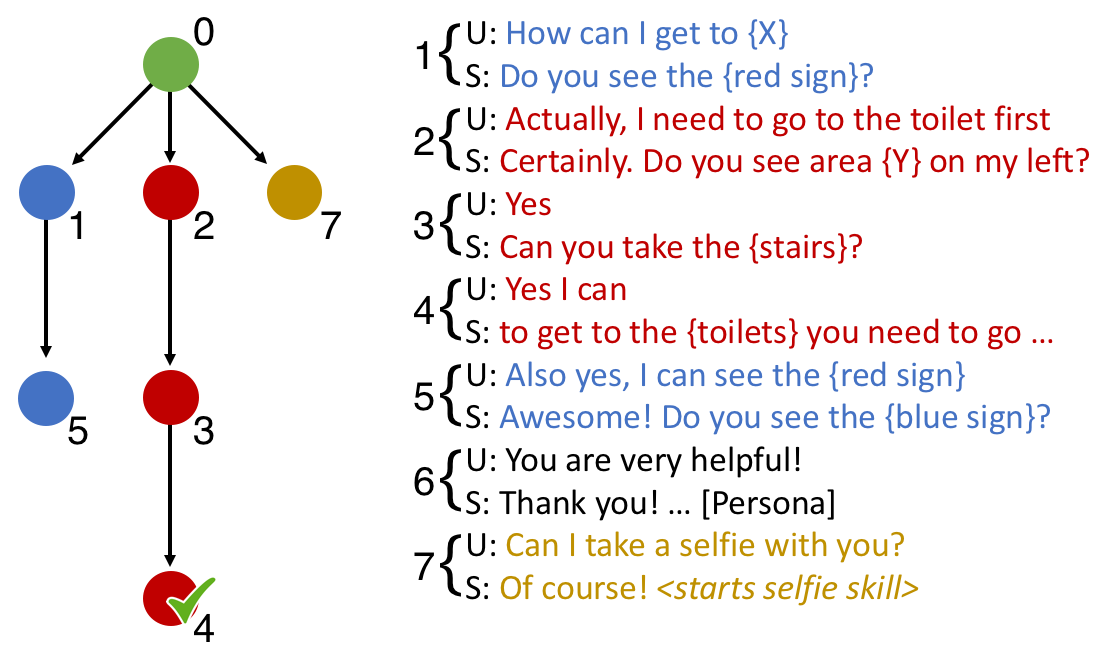}
  \caption{Hypothetical multi-threaded example interaction. Conversation between the User (U) and the System (S). The colours and numbers of the nodes in the tree correspond to the colours and numbers of the text. Each node in the tree represents one turn, i.e. one user utterance + one system utterance. The green root is created at the start of the interaction. Only the red branch has finished. The black text comes from the persona chatbot for social interaction.}
  \label{fig:dialogue}
\end{figure}

\subsection{Concurrent Multi-threaded Execution}
\label{sec:concurrent}

In addition to being able to concurrently execute actions within a \gls{pnm}, the system also allows to concurrently execute several \glspl{pnm} which is a novel contribution compared to other \gls{pn}-based systems. Thinking of the example of dialogue, this might happen if the user asks the system to fulfil a specific task. Within that task, a \texttt{rpn\_action} requires feedback from the user and asks a question. The user however does not answer the question but starts a different task or the same task with other parameters. This means that both plans are executed at the same time. This requires a system that keeps track of the running \glspl{pnm} and assign the answer to a question to the correct \texttt{rpn\_action}. This arbitration system has been introduced in our previous work~\cite{dondrup2017introducing} (see \nameref{sec:experiment} section and Figure~\ref{fig:dialogue}).

\section{Experiment}
\label{sec:experiment}

The work presented in this paper is part of the MuMMER project\footnote{\url{http://www.mummer-project.eu/}} 
with the aim of putting an entertaining and helpful robot in a shopping mall. Hence, we conducted preliminary tests with the system described here in a shopping mall in Finland\footnote{\url{https://lempaala.ideapark.fi}}. As mentioned in the \nameref{sec:intro} section, the main purpose of the described system is to act as an action manager that is able to combine dialogue actions and physical actions in the same planning domain. We previously argued why it would be beneficial to combine the two \cite{papaioannou2017hybrid} and, create a system able of social interaction via chat and the execution of physical tasks on a robot. Hence, the hypotheses of the experiment were i) we are able to combine both dialogue and physical actions in the same plan and domain and ii) the system is able to interleave strands of conversation by pausing tasks while preserving the ability to resume the task where it was left off (see Figure~\ref{fig:dialogue}).

The social component or `social-chat' of the dialogue comes from the system described in \cite{papaioannou2017alana,curryalana} and the physical task execution is handled by the \gls{pnm} described here. The physical tasks provided were direction giving, making the robot dance, and taking a selfie with the robot. For the direction giving the \gls{pnm} coordinated several actions that created the route description, clarified abilities of the user, e.g. if they are able to take stairs or see certain landmarks, and confirmed if the descriptions have been understood. If they were not, it offered to repeat them or directed them to a human for more information. For the other two tasks, the \gls{pnm} triggers built in behaviours of the robot. 

\begin{figure}[t]
  \centering
  \includegraphics[width=.7\linewidth]{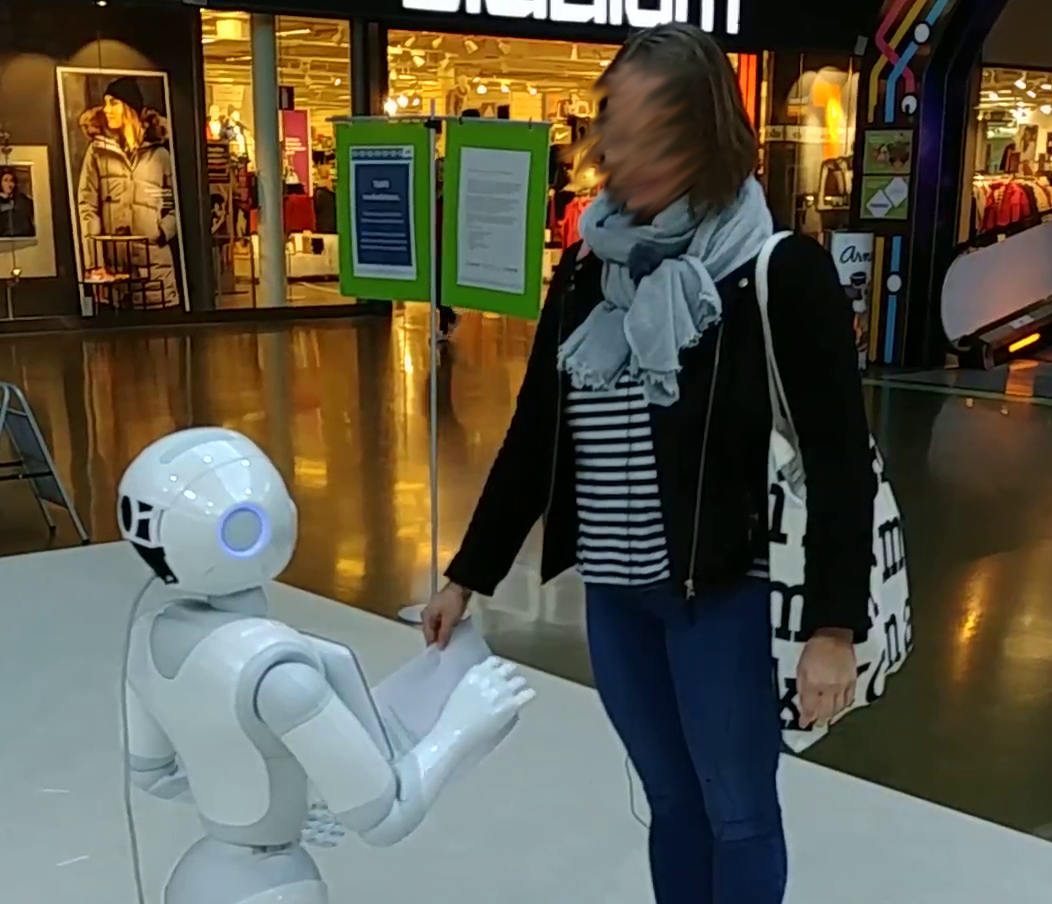}
  \caption{Visitor of the shopping mall interacting with the system embodied by a pepper robot.}
  \label{fig:experiment}
\end{figure}

The robot was deployed in the shopping mall for 5 days during which we tested several different scenarios that were all using \gls{pnm} based execution. These tests included interactions with experimenters but also with customers of the shopping mall (see Figure~\ref{fig:ideapark_me} and \ref{fig:experiment}). We tested scenarios such as interleaving several tasks and interleaving tasks with social chat (see below).

\paragraph{Method and Results}
In order to test our hypotheses, we gave participants a script to follow which does not state what they are supposed to say but rather gives them a list of actions and in which order they should be executed, i.e. you want to find \texttt{shop\_0}, interrupt the robot and ask for \texttt{shop\_1}, interrupt the robot and talk about an unrelated topic such as your favourite artist, film, or book. This interaction was meant to test both hypotheses i) by triggering the guiding task that combines physical (e.g. pointing) and dialogue actions (e.g. describing the route and confirming it has been understood), and ii) that a task can be paused via interrupting its execution with chat or a new task and resume it when being re-prompted by the system because the task has not finished yet. A possible example can be seen in Figure~\ref{fig:dialogue}. This was tested amongst the developers and with 5 customers of the mall who agreed to take part in the experiment. No participation reward was offered.

During these experiments, the customers were rating the system based on the quality of the conversation and the given route descriptions. Since these ratings only mildly reflect the workings of the execution framework, these results have been omitted here. However, through observation we found that in 100\% of the test cases the system triggered the right task, was able to pause it to be interleaved with chat or another task, and finally resumed it when the user was re-prompted with a previous question regarding the task.

\section{Conclusion}

The \gls{pnm} presented here allows developers to automatically create state machines form a simple mark up text file. It uses the modelling powers of \glspl{pn} to handle concurrency and also allows to implement constructs such as loops. In comparison to previous approaches using \glspl{pn} for similar tasks, the \gls{pnm} presented here implements the full functionality of the Petri Net, allows for concurrent execution of several \glspl{pn}, and natively supports \gls{ros} action servers while also allowing to create actions that can interact with a common \gls{kb} during execution. The mark up language used and the underlying implementation gives the user more freedom than previous implementations of this concept as it is more explicit and versatile than a semi-colon separated list of actions (as used in e.g. \cite{ziparo2011petri}) by building on familiar concepts known from languages such as \gls{pddl}. \gls{pddl} style syntax is used for the domain file of the \gls{pnm} while the plan file follows its own unique syntax. However, in comparison to \gls{pddl}, the mark up language of choice, i.e. \gls{yaml}, can be parsed by a wide variety of languages without any programming overhead. 

The system was tested in a shopping mall in Finland as part of a greater dialogue system. 
In this paper we present observational and anecdotal evidence for the \gls{pnm} working as intended. This evidence stems from user tests in the shopping mall that evaluated different parts of a greater system but all used \glspl{pnm} as the underlying execution system.

In conclusion, we have presented a system that is able to generate \glspl{dfsm} as \glspl{pnm}, execute tasks successfully even if concurrent and also execute several \glspl{pnm} concurrently. It presents an alternative to other approaches such as \gls{smach}~\cite{bohren2010smach} or \gls{pnp}~\cite{ziparo2011petri} and provides benefits over each of them.

Future work will investigate the ease of use of the developed modelling language, and the framework itself and its modelling capabilities.

\section{Acknowledgments}
The research leading to these results has received funding from the European Commission’s H2020 programme under grant agreement No.\ 688147, MuMMER project.

\bibliographystyle{aaai}
\bibliography{refs}

\end{document}